\documentclass[10pt,twocolumn,letterpaper]{article}

\usepackage{cvpr}
\usepackage{times}
\usepackage{epsfig}
\usepackage{graphicx}
\usepackage{amsmath}
\usepackage{amssymb}
\usepackage{epigraph}

\usepackage{caption}
\usepackage{xcolor}

\DeclareMathOperator*{\argmin}{arg\,min}

\usepackage[pagebackref=true,breaklinks=true,letterpaper=true,colorlinks,bookmarks=false]{hyperref}

\cvprfinalcopy 

\ifcvprfinal\fi
\begin{document}

\title{Photo Wake-Up: 3D Character Animation from a Single Photo}

\author{Chung-Yi Weng$^1$, \quad
Brian Curless$^1$, \quad
Ira Kemelmacher-Shlizerman$^{1,2}$ \\
$^1$University of Washington\\
$^2$Facebook Inc.
}

\twocolumn[{%
\renewcommand\twocolumn[1][]{#1}%
\maketitle
    \begin{center}
    \includegraphics[width=\textwidth]{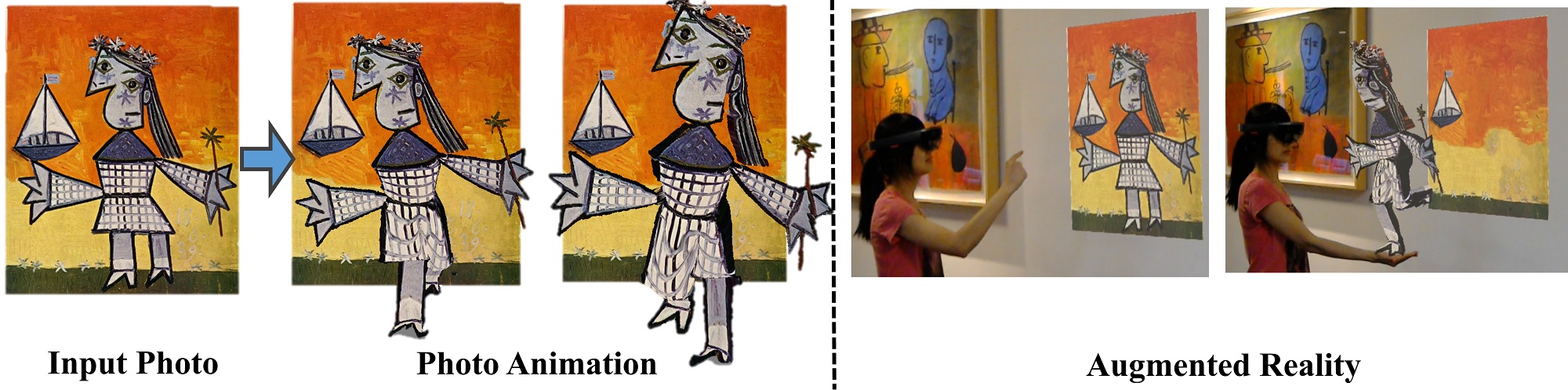}
    \end{center}    
    \captionof{figure}{Given a single photo as input (far left), we create a 3D animatable version of the  subject, which can now walk towards the viewer (middle). The 3D result can be experienced in augmented reality (right); in the result above the user has virtually hung the artwork with a HoloLens headset and can watch the character run out of the painting from different views. \textbf{Please see all results in the supplementary video: }{\small \url{https://youtu.be/G63goXc5MyU}}.\\}
    \label{fig:teaser}
}]


\begin{abstract}
We present a method and application for animating a human subject from a single photo. E.g., the character can walk out, run, sit, or jump in 3D. The key contributions of this paper are: 1) an application of viewing and animating humans in single photos in 3D, 2) a novel 2D warping method to deform a posable template body model to fit the person's complex silhouette to create an animatable mesh, and 3) a method for handling partial self occlusions. We compare to state-of-the-art related methods and evaluate results with human studies. Further, we present an interactive interface that allows re-posing the person in 3D, and an augmented reality setup where the animated 3D person can emerge from the photo into the real world.  We demonstrate the method on photos, posters, and art. The project page is at {\small \url{https://grail.cs.washington.edu/projects/wakeup/}}.

\end{abstract}


\section{Introduction}

\epigraph{Whether you come back by page or by the big screen, Hogwarts will always be there to welcome you home.}{\textit{J.K. Rowling}}

In this paper, we propose to ``wake up a photo'' by bringing the foreground character to life, so that it can be animated in 3D and emerge from the photo.  Related to our application are 
cinemagraphs and GIFs\footnote{Artistic cinemagraphs: \url{http://cinemagraphs.com/}} where a small motion is introduced to a photo to visualize dominant dynamic areas. Unlike a cinemagraph, which is a 2D experience created from video, our method takes a single photo as input and results in a fully 3D experience. The output animation can be played as a video, viewed interactively on a monitor, and as an augmented or virtual reality experience, where a user with an  headset can enjoy the central figure of a photo coming out into the real world.

A central challenge in delivering a compelling experience is to have the reconstructed subject closely match the silhouette of the clothed person in the photo, including self-occlusion of, e.g., the subject's arm against the torso.  Our approach begins with existing methods for segmenting a person from an image, 2D skeleton estimation, and fitting a (semi-nude) morphable, posable 3D model.  The result of this first stage, while animatable, does not conform to the silhouette and does not look natural.

Our key technical contribution, then, is a method for constructing an animatable 3D model that matches the silhouette in a single photo and handles self-occlusion.  Rather than deforming the 3D mesh from the first stage -- a difficult problem for intricate regions such as fingers and for scenarios like abstract artwork -- we map the problem to 2D, perform a silhouette-aligning warp in image space, and then lift the result back into 3D.  This 2D warping approach works well for handling complex silhouettes.  Further, by introducing label maps that delineate the boundaries between body parts, we extend our method to handle certain self-occlusions.

Our operating range on input and output is as follows.  The person should be shown in whole (full body photo) as a fairly frontal view.  We support partial occlusion, specifically of arms in front of the body. While we aim for a mesh that is sufficient for convincing animation, we do not guarantee a metrically correct 3D mesh, due to the inherent ambiguity in reconstructing a 3D model from 2D input.  Finally, as existing methods for automatic detection, segmentation, and skeleton fitting are not yet fully reliable (esp. for abstract artwork), and hallucinating the appearance of the back of a person is an open research problem, we provide a user interface so that a small amount of input can correct errors and guide texturing when needed or desired.

To the best of our knowledge, our system is the first to enable 3D animation of a clothed subject from a single image.  The closest related work either does not recover fully 3D models~\cite{hornung2007character} or is built on video, i.e., multi-view, input~\cite{alldieck2018video}.  We compare to these prior approaches, and finally show results for a wide variety of examples as 3D animations and AR experiences.
\section{Related Work}
\label{sec:related}

General animation from video has led to many creative effects over the years.  The seminal ``Video Textures'' \cite{schodl2000video} work shows how to create a video of infinite length starting from a single video. Human-specific video textures were produced from motion capture videos via motion graphs \cite{flagg2009human}. \cite{xu2011video} explore multi-view captures for human motion animation, and \cite{zhou2012image} demonstrate that clothing can be deformed in user videos guided by  body skeleton and videos of models wearing the same clothing. Cinemagraphs \cite{tompkin2011towards,bai2013automatic} or Cliplets \cite{joshi2012cliplets} create a still with small motion in some part of the still, by segmenting part of a given video in time and space. 

Relevant also are animations created from big data sets of images, e.g., personal photo collections of a person where the animation shows a transformation of a face through years \cite{kemelmacher2011exploring}, or Internet photos to animate transformation of a location in the world through years \cite{martin2015time}, e.g., how flowers grow on Lombard street in San Francisco, or the change of glaciers over a decade. 

Animating from a single photo, rather than videos or photo collections, also resulted in fascinating effects.   \cite{chuang2005animating} animate segmented regions to create an effect of water ripples or swaying flowers. \cite{xu2008animating} predict motion cycles of animals from a still photo of a group of animals, e.g., a group of birds where each bird has a different wing pose.  \cite{kholgade20143d} show that it's possible to modify the 3D viewpoint of  an object in a still by matching to a database of 3D shapes, e.g., rotating a car on in a street photo.  \cite{elor2017bringingPortraits} showed how to use a video of an actor making facial expressions and moving their head to create a similar motion in a still photo.  Specific to body shapes, \cite{zhou2010parametric} showed that it's possible to change the body weight and height from a single image and in a full video \cite{jain2010moviereshape}. \cite{hornung2007character} presented a user-intensive, as-rigid-as-possible 2D animation of a human character in a photo, while ours is 3D. 

For 3D body shape estimation from single photo, \cite{Bogo:ECCV:2016} provided the SMPL model which captures diverse body shapes and proved highly useful for 3D pose and shape estimation applications. Further, using deep networks and the SMPL model, \cite{varol2017learning, kanazawa2017end, pavlakos2018learning} present end-to-end frameworks for single view body pose and shape estimation. \cite{varol18_bodynet} directly infer a volumetric body shape. \cite{Guler2018DensePose} finds dense correspondence between human subjects and UV texture maps. For multi-view, \cite{lun20173d} uses two views (frontal and side) to reconstruct a 3D mesh from sketches. \cite{alldieck2018video} applied SMPL model fitting to video taken while walking around a stationary human subject in a neutral pose to obtain a 3D model, including mesh deformation to approximately fit clothing. Recently, the idea of parametric model has further been extended from humans to animals~\cite{Zuffi:CVPR:2018, cmrKanazawa18}.

Most single-image person animation has focused on primarily 2D or pseudo-3D animation (e.g., \cite{hornung2007character}) while we aim to provide a fully 3D experience.  Most methods for 3D body shape estimation focus on semi-nude body reconstruction and not necessarily ready for animation, while we take cloth into account and look for an animatable solution. The most similar 3D reconstruction work is \cite{alldieck2018video} although they take a video as input. We compare our results to \cite{hornung2007character} and \cite{alldieck2018video} in Sec.~\ref{sec:results}.

\begin{figure*}
  \begin{center}
  \includegraphics[width=\textwidth]{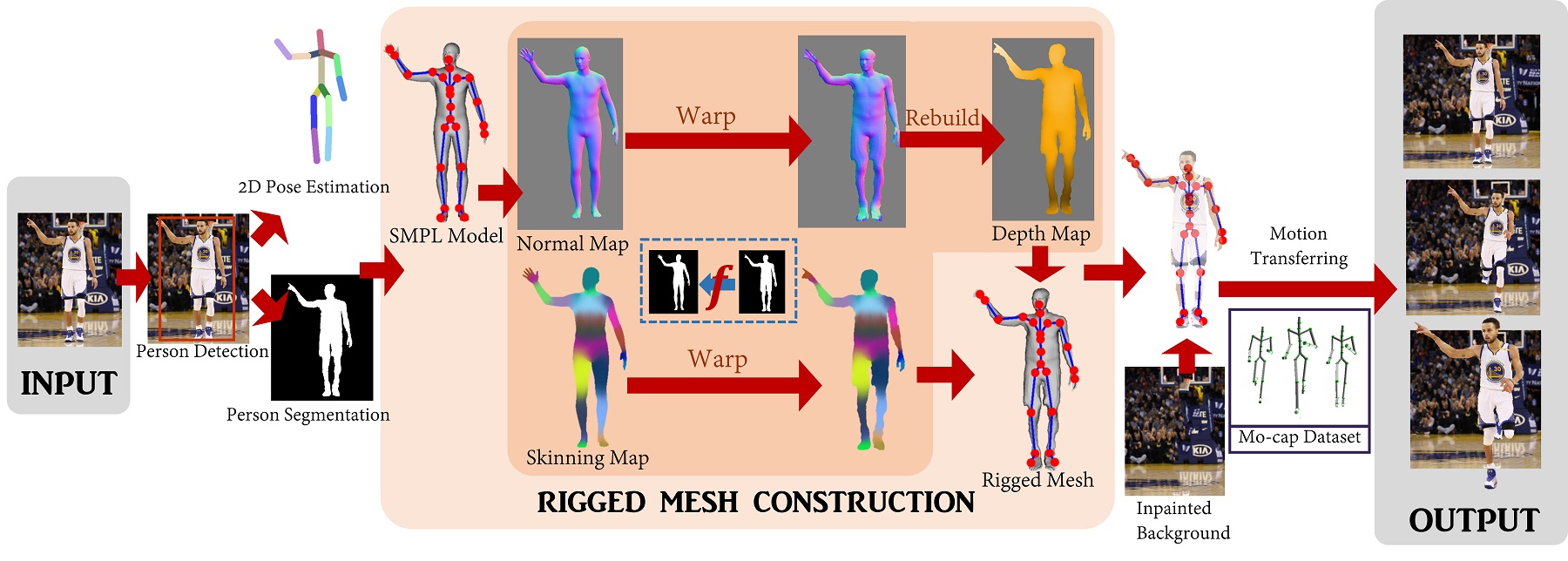}
  \end{center}
  \caption{Overview of our method. Given a photo,  person detection, 2D pose estimation, and person segmentation, is performed using off-the-shelf algorithms. Then, A SMPL template model is fit to the 2D pose and projected into the image as a normal map and a skinning map. The core of our system is:  find a mapping between person's silhouette and the SMPL silhouette,  warp the SMPL normal/skinning maps to the output, and build a depth map by integrating the warped normal map.  This process is repeated to simulate the model's back view and combine depth and skinning maps to create a complete, rigged 3D mesh. The mesh is further textured, and animated using motion capture sequences on an inpainted background.}
  \label{fig:overview}
\end{figure*}

\section{Overview}
\label{sec:overview}

Given a single photo, we propose to animate the human subject in the photo. The overall system works as follows (Fig. \ref{fig:overview}): We first apply state-of-the-art algorithms to perform person detection, segmentation, and 2D pose estimation. From the results, we devise a method to construct a rigged mesh (Section \ref{sec:rigged_mesh_construction}).  Any 3D motion sequence can then be used to animate the rigged mesh.

To be more specific, we use Mask R-CNN \cite{he2017mask} for person detection and segmentation (implementation by \cite{matterport:2017}). 2D body pose is estimated using \cite{wei2016convolutional}, and  person segmentation is refined using Dense CRF \cite{krahenbuhl2011efficient}. Once the person is segmented out of the photo, we apply PatchMatch \cite{barnes2009patchmatch} to fill in the regions where the person used to be. 
\section{Mesh Construction and Rigging}
\label{sec:rigged_mesh_construction}

The key technical idea of this paper is how to recover an animatable, textured 3D mesh from a single photo to fit the proposed application. 

We begin by fitting the SMPL morphable body model~\cite{SMPL:2015} to a photo, including the follow-on method for fitting a shape in 3D to the 2D skeleton~\cite{Bogo:ECCV:2016}. The recovered SMPL model provides an excellent starting point, but it is semi-nude, does not conform to the underlying body shape of the person and, importantly, does not match the clothed silhouette of the person.  

One way is to force the SMPL model to fit the silhouettes by optimizing vertex locations on the SMPL mesh, taking care to respect silhouette boundaries, avoid pinching, and  self-intersection.  This is challenging especially around intricate regions such as fingers. This was indeed explored by \cite{alldieck2018video}, and we compare to those results in the experiments.

Instead, we take a 2D approach: warp the SMPL silhouette to match the person silhouette in the original image and then apply that warp to projected SMPL normal maps and skinning maps. The resulting normal and skinning maps can be constructed for both front and (imputed) back views and then lifted into 3D, along with the fitted 3D skeleton, to recover a rigged body mesh that exactly agrees with the silhouettes, ready for animation.  The center box in Figure~\ref{fig:overview} illustrates our approach.

In the following, we describe how we construct a rigged mesh using 2D warping (Section \ref{sec:mesh_warping}), then present how to handle arm-over-body self-occlusion (Section \ref{sec:deal_with_self-occlusion}).

\subsection{Mesh Warping, Rigging, \& Skinning}
\label{sec:mesh_warping}
In this section, we describe the process for constructing a rigged mesh for a subject without self-occlusion.

We start with the 2D pose of the person and the person's silhouette mask $S$.  For simplicity, we refer to $S$ both as a set and as a function, i.e., as the set of all pixels within the silhouette, and as a binary function $S(x)=1$ for pixel $x$ inside the silhouette or $S(x)=0$ for $x$ outside the silhouette.

To construct a 3D mesh with skeletal rigging, we first fit a SMPL model to the 2D input pose using the method proposed by~\cite{Bogo:ECCV:2016}, which additionally recovers camera parameters.  We then project this mesh into the camera view to form a silhouette mask $S_{\rm SMPL}$.  The projection additionally gives us a depth map $Z_{\rm SMPL}(x)$, a normal map $N_{\rm SMPL}(x)$ and a skinning map $W_{\rm SMPL}(x)$ for pixels $x \in S_{\rm SMPL}$. The skinning map is derived from the per-vertex skinning weights in the SMPL model and is thus vector-valued at each pixel (one skinning weight per bone).

Guided by $S_{\rm SMPL}$ and the input photo's silhouette mask $S$, we then warp $Z_{\rm SMPL}$, $N_{\rm SMPL}$, and $W_{\rm SMPL}$ to construct an output depth map (at the silhouette only) $Z_{\partial S}(x \in  \partial S)$, normal map $N(x)$, and skinning map $W(x)$, respectively, for pixels $x \in S$.  $N(x)$ is then integrated to recover the final depth map $Z(x)$, subject to matching $Z_{\partial S}(x)$ at the silhouette boundary $\partial S$. More concretely, we solve for a smooth inverse warp, $f(x)$, such that:
\begin{equation}
S(x) = S_{\rm SMPL}(f(x))
\end{equation}
and then apply this warp to the depth and skinning maps:
\begin{align}
Z_{\partial S}(x \in \partial S)& = Z_{\rm SMPL}(f(x)) \\
N(x)& = N_{\rm SMPL}(f(x)) \\
Z(x)& = {\rm Integrate}[N; Z_{\partial S}]\\
W(x)& = W_{\rm SMPL}(f(x))
\end{align}

\begin{figure}[ht]
  \centering
  \includegraphics[width=1.0\linewidth]{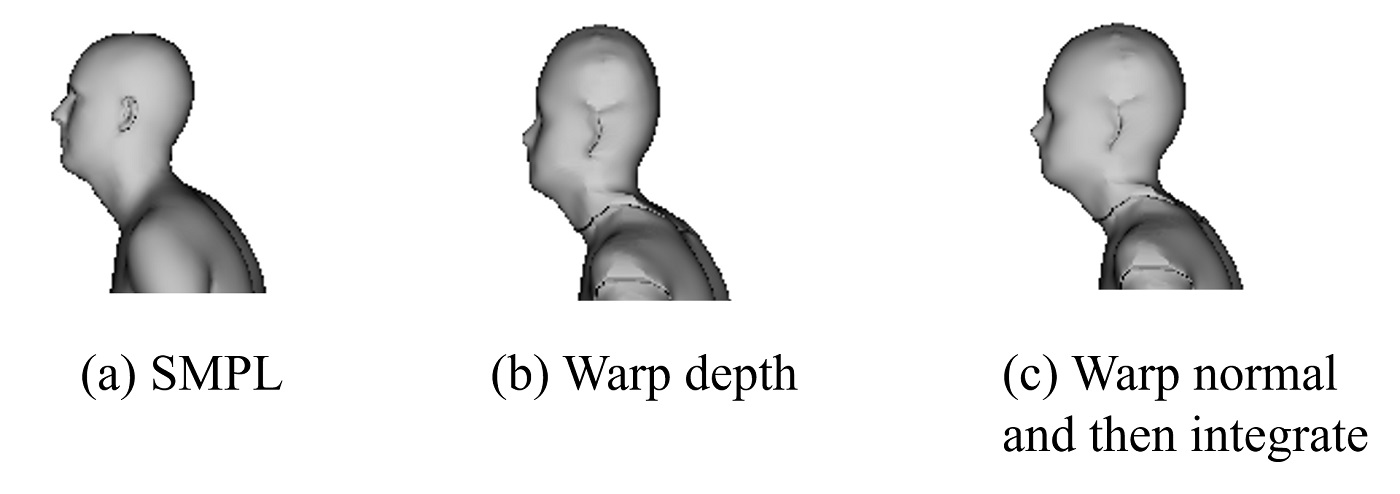}
  \caption{Comparison of different depth map constructions, after stitching front and back depth maps together (Section~\ref{sec:mesh_rebuilding}). Given (a) a reference SMPL model, we can reconstruct a mesh (b) by warping the SMPL depth maps or (c) by warping the SMPL normal maps and then integrating. Notice the flattening evident in (b), particularly around the head.}
  \label{fig:normal_comparison}
\end{figure}

We experimented with setting $Z(x) = Z_{\rm SMPL}(f(x))$, but the resulting meshes were usually too flat in the $z$ direction (See Fig. \ref{fig:normal_comparison}b). The warping procedure typically stretches the geometry in the plane (the SMPL model is usually thinner than the clothed subject, often thinner than even the unclothed subject), without similarly stretching (typically inflating) the depth.  We address this problem by instead warping the {\em normals} to arrive at $N(x)$ and then integrating them to produce $Z(x)$.  In particular, following \cite{basri2007photometric}, we solve a sparse linear system to produce a $Z(x)$ that agrees closely with the warped normals $N(x)$ subject to the boundary constraint that $Z(x) = Z_{\partial S}(x)$ for pixels $x \in \partial S$. Fig. \ref{fig:normal_comparison} shows the difference between the two methods we experimented with. 

To construct the inverse warp, $f(x)$, many smooth warping functions are possible; we choose one based on mean-value coordinates~\cite{floater2003mean} because it is well defined over the entire plane for arbitrary planar polygons without self-intersections, which fits our cases very well. In particular, given the ordered set of points (vertices) on the closed polygonal boundary of the input silhouette, 
$p_i \in \partial{S}=(p_0, p_1, \dots, p_{m-1})$, we can represent any point inside of $S$ as:
\begin{equation}
\label{eqn:generalized_coordinates}
x = \displaystyle\sum_{i=0}^{m-1} \lambda_i(x)p_i
\end{equation}
where $(\lambda_0(x), \lambda_1(x), \dots, \lambda_{m-1}(x))$ are the mean-value coordinates of any $x \in S$ with respect to the boundary vertices $p_i$.  

Suppose we have a correspondence function $\phi$ that identifies $p_i$ on the input silhouette boundary $\partial{S}$ with points on the SMPL silhouette boundary $p^{\rm SMPL}_i \in \partial{S_{\rm SMPL}}=(p^{\rm SMPL}_0, p^{\rm SMPL}_1, \dots, p^{\rm SMPL}_{n-1})$:
\begin{equation}
p_i \rightarrow p^{\rm SMPL}_{\phi[i]}.
\end{equation}
Then, using the same mean-value coordinates from Eq.~\ref{eqn:generalized_coordinates}, we define the warp function to be:
\begin{equation}
f(x) = \displaystyle\sum_{i=0}^{m-1} \lambda_i(x)p^{\rm SMPL}_{\phi[i]}.
\end{equation}
Next, we describe how we compute the correspondence function $\phi$, fill holes in the normal and skinning maps, and then construct a complete mesh with texture.

\subsubsection{Boundary matching}
\label{sec:boundary_matching}
We now seek a mapping $\phi$ that provides correspondence between points $p_i \in \partial{S}$ and points $p^{\rm SMPL}_j \in \partial{S_{\rm SMPL}}$.  We would like each point $p_i$ to be close to its corresponding point $p^{\rm SMPL}_{\phi[i]}$, and, to encourage smoothness, we would like the mapping to be monotonic without large jumps in the indexing.  To this end, we solve for $\phi[i]$ to satisfy:
\begin{equation}
\argmin_{\phi[0], \dots, \phi[m-1]}{\displaystyle\sum_{i=0}^{m-1} D(p_i, p^{\rm SMPL}_{\phi[i]}) + T(\phi[i], \phi([i+1]))}
\end{equation}
where
\begin{equation}
D(p_i, p^{\rm SMPL}_{\phi[i]}) = \| p_i - p^{\rm SMPL}_{\phi[i]} \|_2
\end{equation}
and
\begin{equation}
T(\phi[i], \phi[i+1]) = 
\begin{cases}
	1, & \quad \text{if } 0 \leq \phi[i+1] - \phi[i] \leq \kappa\\
    \infty, & \quad \text{otherwise}
\end{cases}
\end{equation}
$D(p_i, p^{\rm SMPL}_{\phi[i]})$ is designed to encourage closeness of corresponding points, and $T(\phi[i], \phi[i+1])$ avoids generating an out-of-order sequence with big jumps.  Because we are indexing over closed polygons, we actually use $\phi[i \% m] \% n$ in the objective.  With $\kappa = 32$, we  solve for $\phi$  with dynamic programming.

\subsubsection{Hole-filling}

In practice, holes may arise when warping by $f(x)$, i.e., small regions in which $f(x) \notin S_{\rm SMPL}$, due to non-bijective mapping between $\partial{S}$ and $\partial{S_{\rm SMPL}}$. We smoothly fill these holes in the warped normal and skinning weight maps. Please refer to the supplemental material for more detail and illustration of the results of this step.

\subsubsection{Constructing the complete mesh}
\label{sec:mesh_rebuilding}

The method described so far recovers depth and skinning maps for the front of a person.  To recover the back of the person, we virtually render back view of the fitted SMPL model, mirror the person mask, and then apply the warping method described previously.

\begin{figure}[ht]
  \centering
  \includegraphics[width=\linewidth]{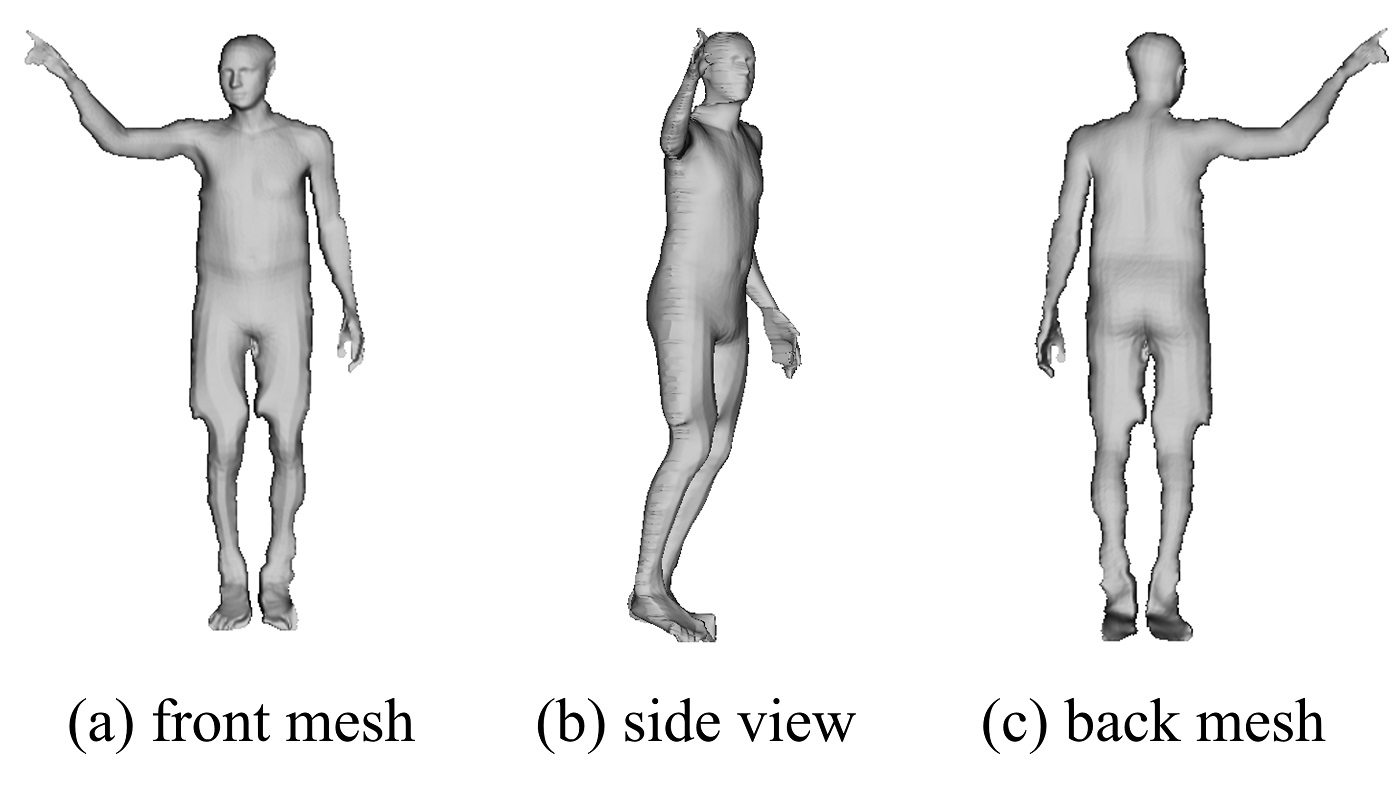}
  \caption{Reconstructed mesh results.  We reconstruct the front mesh (a) and the back mesh (c) separately and then combine them into one mesh, viewed from the side in (b).}
  \label{fig:mesh_results}
\end{figure}

We reconstruct front and back meshes in the standard way: back-project depths into 3D and construct two triangles for each 2x2 neighborhood.  We assign corresponding skinning weights to each vertex.  Stitching the front and back meshes together is straightforward as they correspond at the boundary.  Fig. \ref{fig:mesh_results} illustrates the front and back meshes and the stitched model.

\begin{figure*}
  \centering
  \includegraphics[width=0.88\linewidth]{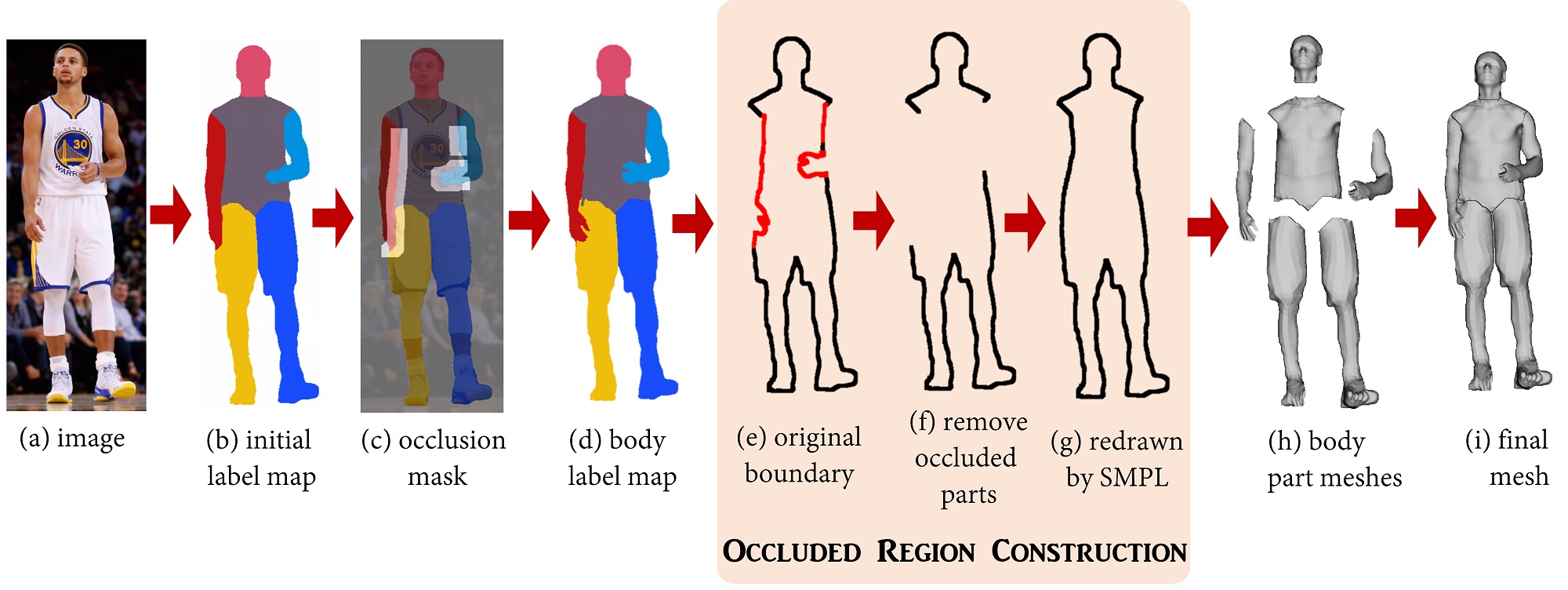}
  \caption{Starting from the input image (a) and its corresponding silhouette and projected SMPL body part model, we recover an initial body part label map (b).  After identifying points at occlusion boundaries, we construction an occlusion mask (lighter areas in (c)) and then refine it to construct the final body label map (d).  The body part regions near occlusions have spurious boundaries, shown in red in (e).  We remove these spurious boundaries (f) and replace them with transformed versions of the SMPL boundaries (g).  We can then reconstruct the body part-by-part (h) and assemble into the final mesh (i).}
  \label{fig:occlusion_overview}
\end{figure*}

\subsection{Self-occlusion}
\label{sec:deal_with_self-occlusion}

When the subject self-occludes -- one body part over another -- reconstructing a single depth map (e.g., for the front) from a binary silhouette will not be sufficient.  To handle self-occlusion, we segment the body into parts via body label map, complete the partially occluded segments, and then reconstruct each part using the method described in Section~\ref{sec:mesh_warping}.  Fig.~\ref{fig:occlusion_overview} illustrates our approach.

We focus on self-occlusion when the arms partially cross other body parts such that the covered parts are each still a single connected component. Our method does not handle all self-occlusion scenarios, but does significantly extend the operating range and show a path toward handling more cases.

\subsubsection{Body label map}
\label{sec:body_label_map}

The projected SMPL model provides a reference body label map $L_{\rm SMPL}$ that does not conform closely to the image.  We use this label map to construct a final label map $L$ in two stages: (1) estimate an initial label map $L_{\rm init}$ for each pixel $x \in S$ to be as similar as possible to $L_{\rm SMPL}$, then (2) refine $L_{\rm init}$ at occlusion boundaries where the label discontinuities should coincide with edges in the input image.

\textbf{Initial Body Labeling.} We solve for the initial (rough) body label map $L_{\rm init}$ by minimizing a Markov Random Field (MRF) objective:
\begin{equation}
\min_{L_{\rm init}} \displaystyle\sum_{p \in S} U(L_{\rm init}(p)) + \; \gamma\!\!\!\!\!\!\!\!\!\!\!\!\displaystyle\sum_{p \in S, q \in \mathcal{N}(p) \cap S} \!\!\!\!\!\!\!\!\!\!\! V(L_{\rm init}(p), L_{\rm init}(q))
\end{equation}
where
\begin{equation}
U(L_{\rm init}(p)) = \min_{r \mid L_{\rm SMPL}(r) = L(p)} \|p-r\|_2
\end{equation}
\begin{equation}
V(L_{\rm init}(p), L_{\rm init}(q)) = 
	\begin{cases}
		1 & \quad \text{if } L_{\rm init}(p) \neq L_{\rm init}(q) \\
        0 & \quad \text{otherwise}
	\end{cases}
\end{equation}
$\mathcal{N}(p)$ is the 8-neighborhood of $p$.  $U(.)$ scores a label according to the distance to the nearest point in $L_{\rm SMPL}$ with the same label, thus encouraging $L_{\rm init}$ to be similar in shape to $L_{\rm SMPL}$, while $V(.)$ encourages spatially coherent labels.

We use $\alpha$-expansion~\cite{boykov2001fast} to approximately solve for $L_{\rm init}$, with $\gamma=16$. Fig.~\ref{fig:occlusion_overview}(b) illustrates the initial label map produced by this step.

\textbf{Refined Body Labeling.} Next, we refine the body label map to more cleanly separate occlusion boundaries. 

Occlusion boundaries occur when two pixels with different part labels are neighbors in the image, but are not neighbors on the 3D body surface.  To identify these pixels, we first compute warp functions $f_{\ell}$ that map each body part $L_{\rm init}=\ell$ to the corresponding body part $L_{\rm SMPL}=\ell$, using the mean-value coordinate approach described in Section~\ref{sec:mesh_warping}, performed part-by-part. Then, along the boundaries of arm parts of $L_{\rm init}$, for each pair of neighboring pixels $(p,q)$ with different labels, we determine the corresponding projected SMPL locations $(f_{L_{\rm init}(p)}(p),f_{L_{\rm init}(q)}(q))$, back-project them onto the SMPL mesh, and check if they are near each other on the surface.  If not, these pixels are identified as occlusion pixels.  Finally, we dilate around these occlusion pixels to generate an occlusion mask $O$. The result is shown in Fig.~\ref{fig:occlusion_overview}(c).

We now refine the labels within $O$ to better follow color discontinuities in the image $I$, giving us the final body label map $L$.  For this, we define another MRF:
\begin{equation}
\min_{L} \displaystyle\sum_{p \in O} U(L(p)) + \:\: \gamma\!\!\!\!\!\!\!\!\!\displaystyle\sum_{p \in O, q \in \mathcal{N}(p)} \!\!\!\!\!\!\!\!\! V(L(p), L(q))
\end{equation}
where
\begin{equation}
	U(L(p)) = -\log({\rm GMM}(L(p), I(p)))
\end{equation}
\begin{equation}
	V(L(p), L(q)) = C(L(p), L(q))e^{-\beta \|I(p)-I(q)\|^2}
\end{equation}
\begin{equation}
	C(L(p), L(q)) =
    \begin{cases}
    	1/\|p-q\| & \quad \text{if } L(p) \neq L(q) \\
        0	   & \quad \text{otherwise}
    \end{cases}
\end{equation}
where ${\rm GMM}(L(p), I(p))$ is the probability of $p$ with color $I(p)$ labeled as $L(p)$, modeled using a Gaussian Mixture Model. We set $\gamma=8$, and, following~\cite{rother2004grabcut}, we set $\beta$ to be:
\begin{equation}
	\beta = (2\langle \|(I(p) - I(q)\|^2 \rangle)^{-1}
\end{equation}
where $\langle . \rangle$ averages over all pairs of neighboring pixels in $O$.  

The problem is solved by iteratively applying $\alpha$-expansions \cite{boykov2001fast}, where in each iteration we re-estimate ${\rm GMM}(.)$ using the latest approximated $L$ initizlied as $L_{\rm init}$. Fig~\ref{fig:occlusion_overview}(d) illustrates the final body map.

\subsubsection{Occluded region construction}
\label{sec:occluded_region_hallucination}

We now have the shapes of the unoccluded segments; the next challenge is to recover the shapes of the partially occluded parts.

We first combine the labels of the head, torso, and legs together into one region $B$. Then we extract the boundary $\partial{B}$ and identify the occlusion boundaries, $\partial{B}^{ocl} \in \partial{B} \cap O$ (shown in red in Fig.~\ref{fig:occlusion_overview}(e)). Next, for a contiguous set of points $\partial{B}^{ocl}_{i} \in \partial{B}^{ocl}$ (e.g., one of the three separate red curves in Fig.~\ref{fig:occlusion_overview}(e)), we find the corresponding boundary $\partial{B_{\rm SMPL}^{ocl}} \in \partial{B_{\rm SMPL}}$ using the the boundary matching algorithm from Section~\ref{sec:boundary_matching}, where $B_{\rm SMPL}$ is the region formed by projecting the SMPL head, torso, and legs to the image plane. We then replace $\partial{B}^{ocl}_{i}$ with $\partial{B_{\rm SMPL}^{ocl}}$ by a similarity transform defined by the end points of $\partial{B}^{ocl}_{i}$ and $\partial{B_{\rm SMPL}^{ocl}}$, as shown in Fig.\ref{fig:occlusion_overview}-(f) and (g).

\subsubsection{Mesh construction}
Once we have completed body labeling and recovered occluded shapes, we project the SMPL model part-by-part to get per-part SMPL depth, normal, and skinning weight maps, then follow the approach in Section~\ref{sec:mesh_warping} to build part meshes (Fig.\ref{fig:occlusion_overview}-(h)), and assemble them together to get our final body mesh (Fig.\ref{fig:occlusion_overview}-(i)). Finally, we apply Laplacian smoothing to reduce jagged artifacts along the mesh boundaries due to binary silhouette segmentation.

\begin{figure*}
  \centering
  \includegraphics[width=\textwidth]{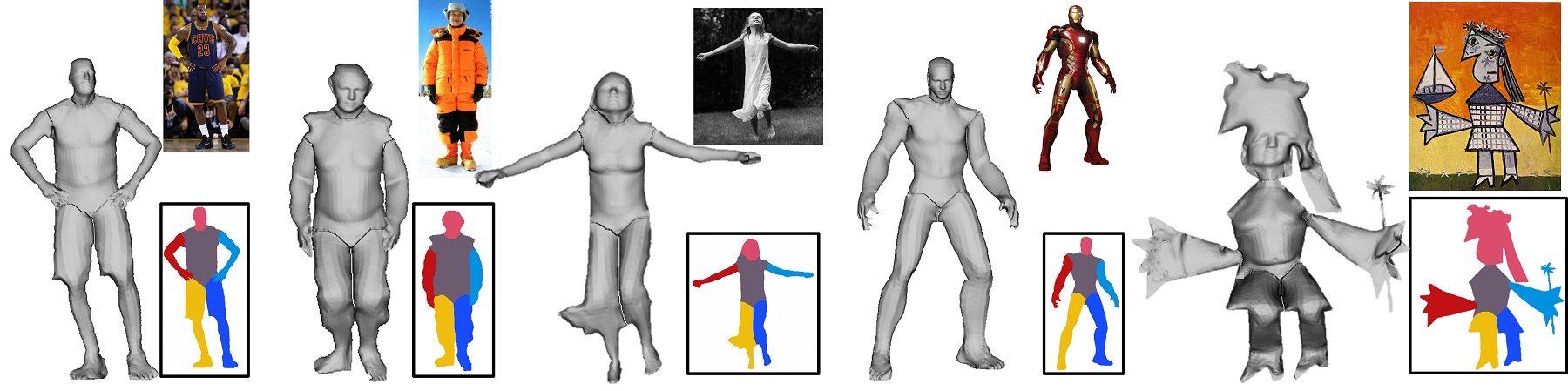}
  \caption{Examples of computed body label maps and meshes (input photos are put on top right corner). }
  \label{fig:more_mesh_restuls}
\end{figure*}

\section{Final Steps}

\textbf{Head pose correction:} Accuracy in head pose is important for good animation, while the SMPL head pose is often incorrect.  Thus, as in \cite{king2009dlib, kemelmacher2011face}, we detect facial fiducials in the image and solve for the 3D head pose that best aligns the corresponding, projected 3D fiducials with the detected ones.  After reconstructing the depth map for the head as before, we apply a smooth warp that exactly aligns the projected 3D fiducials to the image fiducials. Whenever the face or fiducials are not detected, this step is skipped.

\textbf{Texturing:} For the front of the subject, we project the image onto the geometry.  Occluded, frontal body part regions are filled using PatchMatch \cite{barnes2009patchmatch}. Hallucinating the back texture is an open research problem ~\cite{ma2017pose, esser2018variational, ma2017disentangled}.  We provide two options: (1) paste a mirrored copy of the front texture onto the back, (2) inpaint with optional user guidance.  For the second option, inpainting of the back is guided by the body label maps, drawing texture from regions with the same body labels.  The user can easily alter these label maps to, e.g., encourage filling in the back of the head with hair texture rather than face texture. Finally the front and back textures are stitched with poisson blending \cite{perez2003poisson}.

Please refer to the supplemental material for more details and illustrations of head pose correction and texturing.
\section{Results and Discussion}
\label{sec:results}

\begin{figure}[ht]
  \centering
  \includegraphics[width=0.95\linewidth]{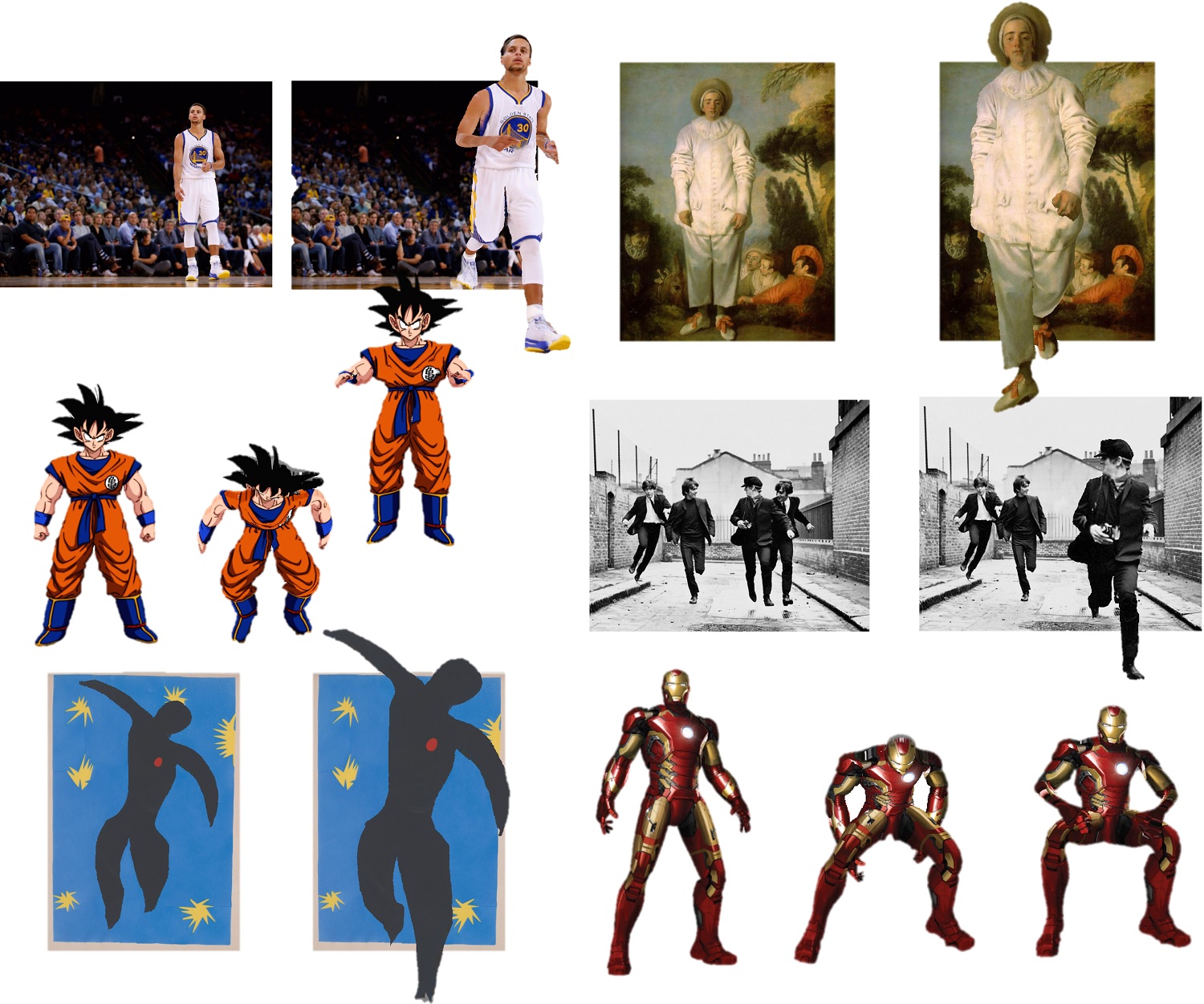}
  \caption{Six animation results. The input is always on left.}
  \label{fig:animaton_result}
\end{figure}

\begin{figure}
  \centering
  \includegraphics[width=0.9\linewidth]{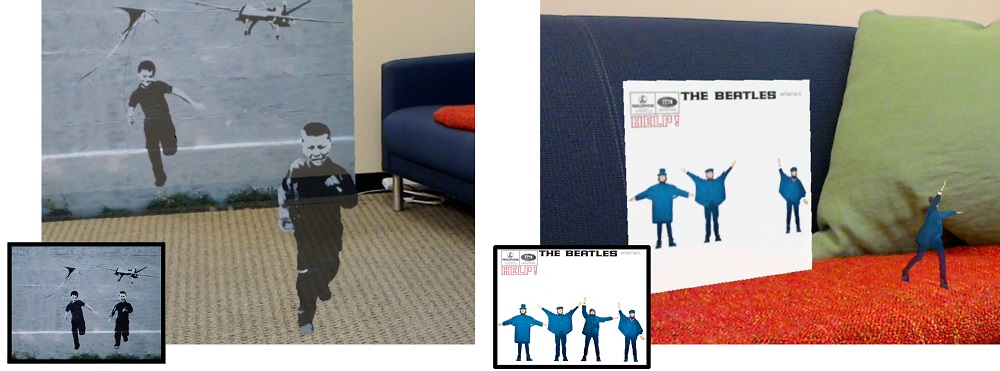}
  \caption{Augmented reality results of our method for different environments (input photos inset). The floor (left) and couch (right) are real, while the people are augmented. }
  \label{fig:ar-result}
\end{figure}

Below we describe our user interface, results, comparisons to related methods, and human study. We have tested our method on  70 photos downloaded from the Internet (spanning art, posters, and graffiti that satisfied our photo specifications--full body, mostly frontal). Figs.~\ref{fig:animaton_result} and \ref{fig:ar-result} show our typical animation and augmented reality results.   With our UI, the user can change the viewpoint during animation, and edit the human pose.  With an AR headset, the user can place the artwork on the wall and walk around the animation while it is playing.  {\bf Please refer to the supplementary video\footnote{\url{https://youtu.be/G63goXc5MyU}} for dynamic versions of the results. }

\textbf{User interface:}  We have created a user interface where the user can interactively: (1) Modify the animation: the default animation is ``running'', where the user can keep some body parts fixed, change the sequence (e.g., choose any sequence from \cite{cmumocap2007}), modify pose and have the model perform an action starting from the modified pose. (2) Improve the automatic detection box, skeleton, segmentation, and body label map if they wish. (3) Choose to use mirrored textures for the back or make adjustments via editing of the body label map. The user interaction time for (2) and (3) is seconds, when needed.

\begin{figure}[ht]
  \centering
  	\includegraphics[width=1.0\linewidth]{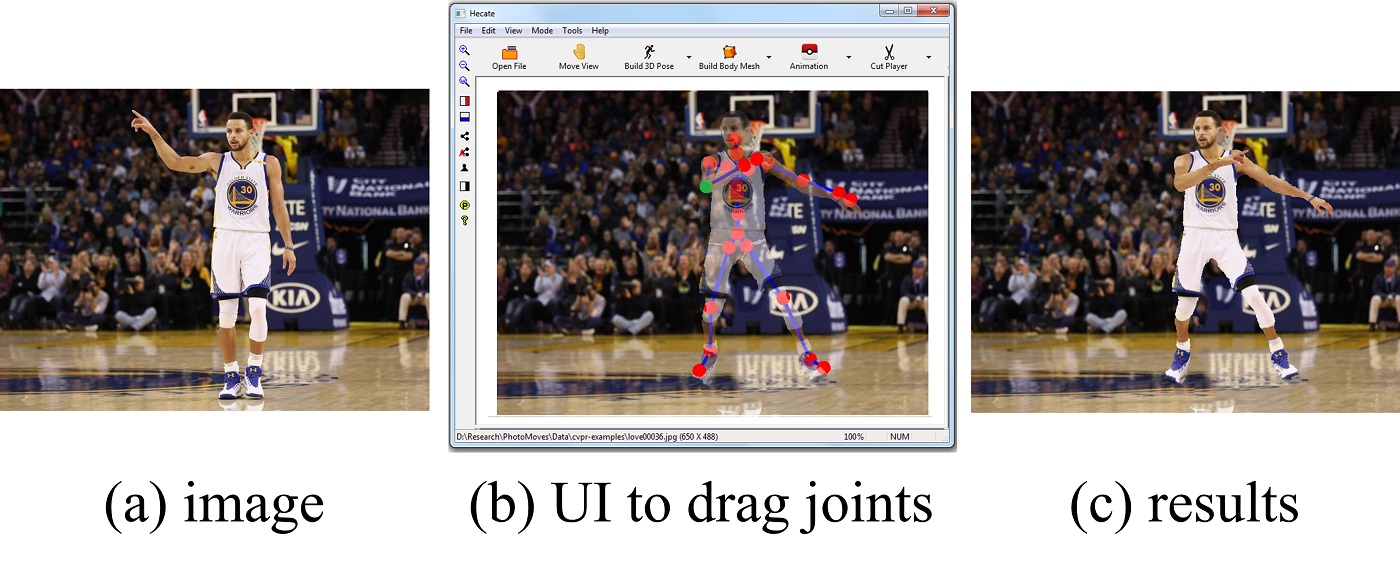}
    \caption{Our user interface for pose editing: (a) Input photo. (b) Editing pose by dragging joints. (c) Result.}
  \label{fig:pose_editing}
\end{figure}

Fig. \ref{fig:pose_editing} shows an example of the pose editing process.  In our UI the mesh becomes transparent to reveal the body skeleton.  By selecting and dragging the joints the user can change the orientation of the corresponding bones. A new image where the pose is edited can be then easily generated. 

The underlying reconstructed geometry for several examples is shown in Fig.~\ref{fig:more_mesh_restuls}. The resulting meshes do not necessarily represent the exact 3D geometry of the underlying subject, but they are sufficient for animation in this application and outperform state of the art as shown below.

\textbf{Human study:} We  evaluated our animation quality via Amazon Mechanical Turk. We tested all 70 examples we produced, rendered as videos. Each result was evaluated by 5 participants, on a scale of 1-3 (1 is bad, 2 is ok, 3 is `nice!'). 350 people responded, and the average score was 2.76 with distribution: 1: 0.6\%, 2: 22.0\%, 3: 77.4\%.

\begin{figure}[ht]
  \centering
  \includegraphics[width=\linewidth]{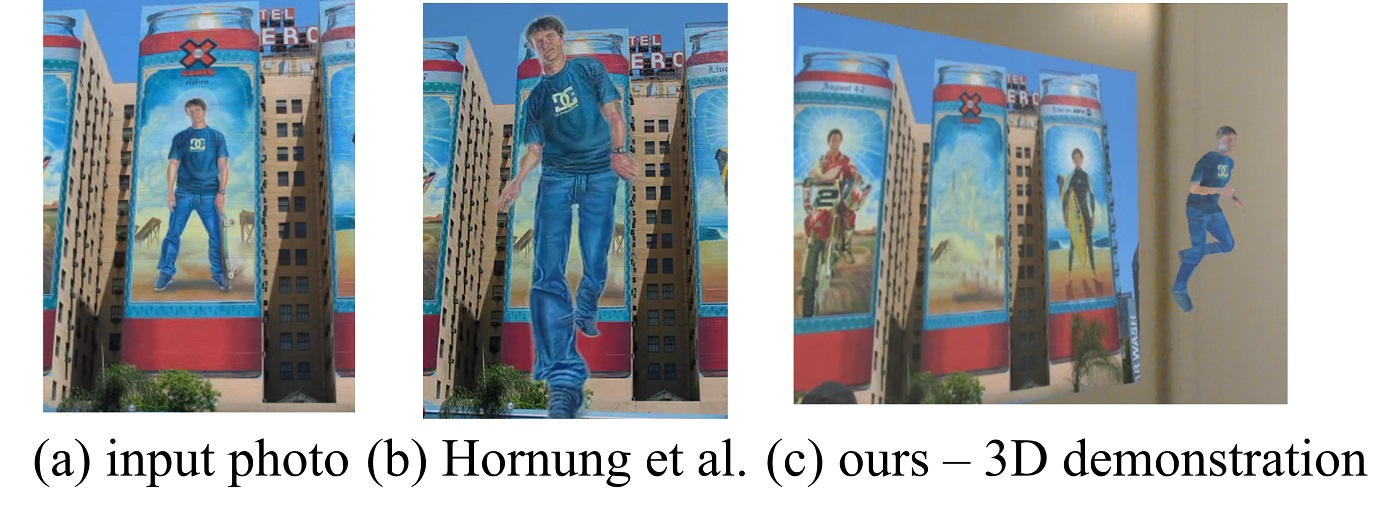}
  \caption{Comparison result with \cite{hornung2007character}: (a) input photo; (b) animation method proposed in \cite{hornung2007character};
  (c) 3D demonstration using our method, which is not possible in~\cite{hornung2007character}.}
  \label{fig:compare_with_hornung}
\end{figure}

\textbf{Comparison with \cite{hornung2007character}:}
We have run our method on the only example in \cite{hornung2007character} that demonstrated substantial out-of-plane motion rather than primarily in-plane 2D motion (see Fig.~\ref{fig:compare_with_hornung}). Our result appears much less distorted in still frames (due to actual 3D modeling) and enables 3D experiences (e.g., AR) not possible in \cite{hornung2007character}. We verified our qualitative observation with a user study on MTurk, asking users to decide which animation is ``more realistic.'' 103 participants responded, and 86\% preferred ours.

\begin{figure}[ht]
  \centering
  \includegraphics[width=\linewidth]{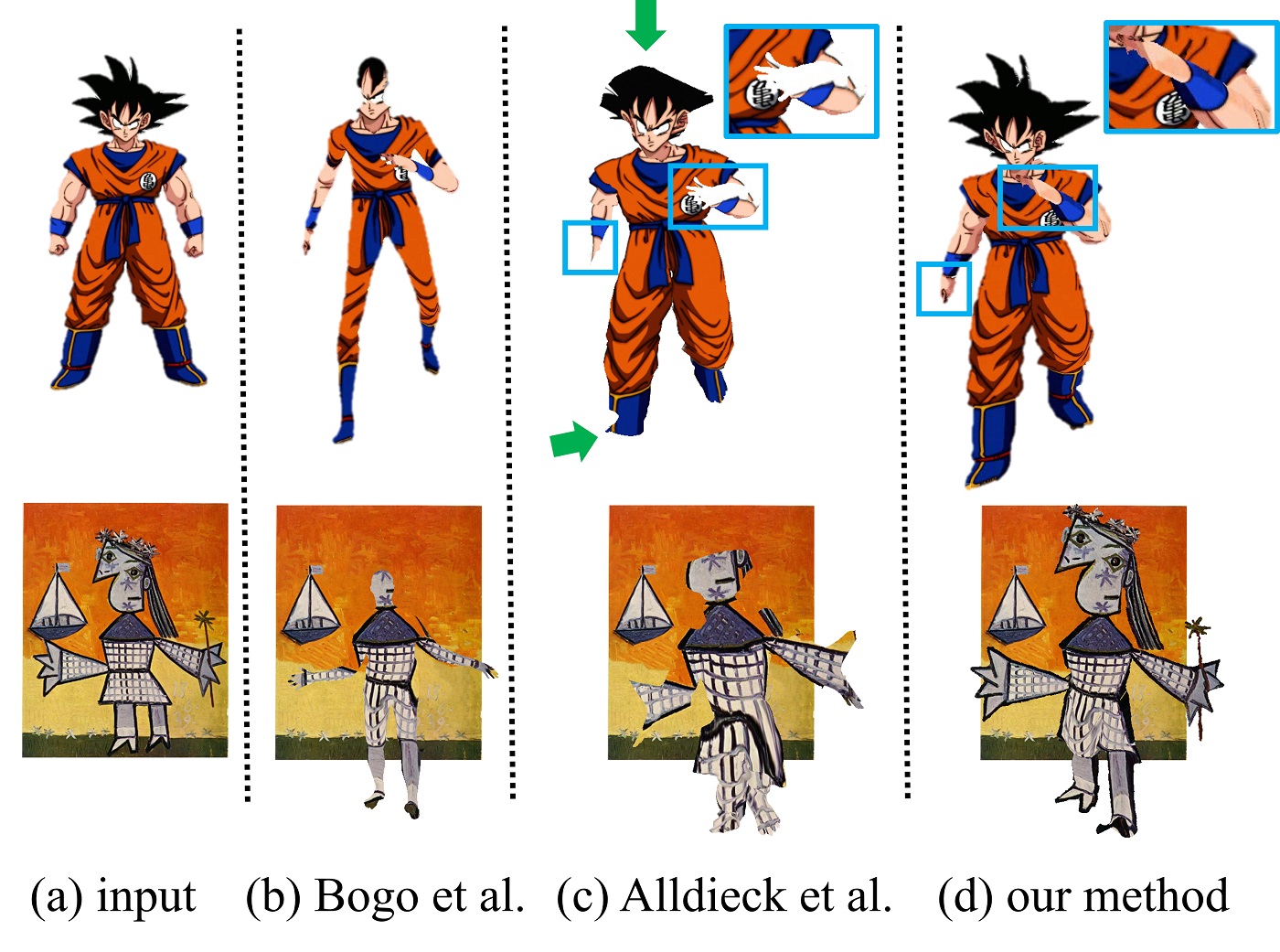}
  \caption{ Comparison with \cite{Bogo:ECCV:2016, alldieck2018video}. (a) Input photo. (b) A fitted SMPL model \cite{Bogo:ECCV:2016}. (c) A deformed mesh using \cite{alldieck2018video}. Notice the mesh does not fit the silhouette (green arrow) and fails to deform fingers (blue box). (d) Our mesh. }
  \label{fig:compare_with_smpl}
\end{figure}

\textbf{Comparison with Bogo et al. \cite{Bogo:ECCV:2016}:} As shown in Fig.~\ref{fig:compare_with_smpl}(b), the fitted, semi-nude SMPL model \cite{Bogo:ECCV:2016} does not correctly handle subject silhouettes.

\textbf{Comparison with Alldieck et al. \cite{alldieck2018video}}: 
In \cite{alldieck2018video}, a SMPL mesh is optimized to approximately match silhouettes of a static subject in a multi-view video sequence.  Their posted code uses 120 input frames, with objective weights tuned accordingly; we thus provide their method with 120 copies of the input image, in addition to the same 2D person pose and segmentation we use.  The results are shown in Fig.~\ref{fig:compare_with_smpl}(c). Their method does not fit the silhouette well; e.g., smooth SMPL parts don't become complex (bald head mapped to big, jagged hair) and the detailed fingers are not warped well to the closed fists or abstract art hands.  Further, it does not handle self-occlusion well, since the single-image silhouette does not encode self-occlusion boundaries.

\begin{figure}[ht]
  \centering
  \includegraphics[width=\linewidth]{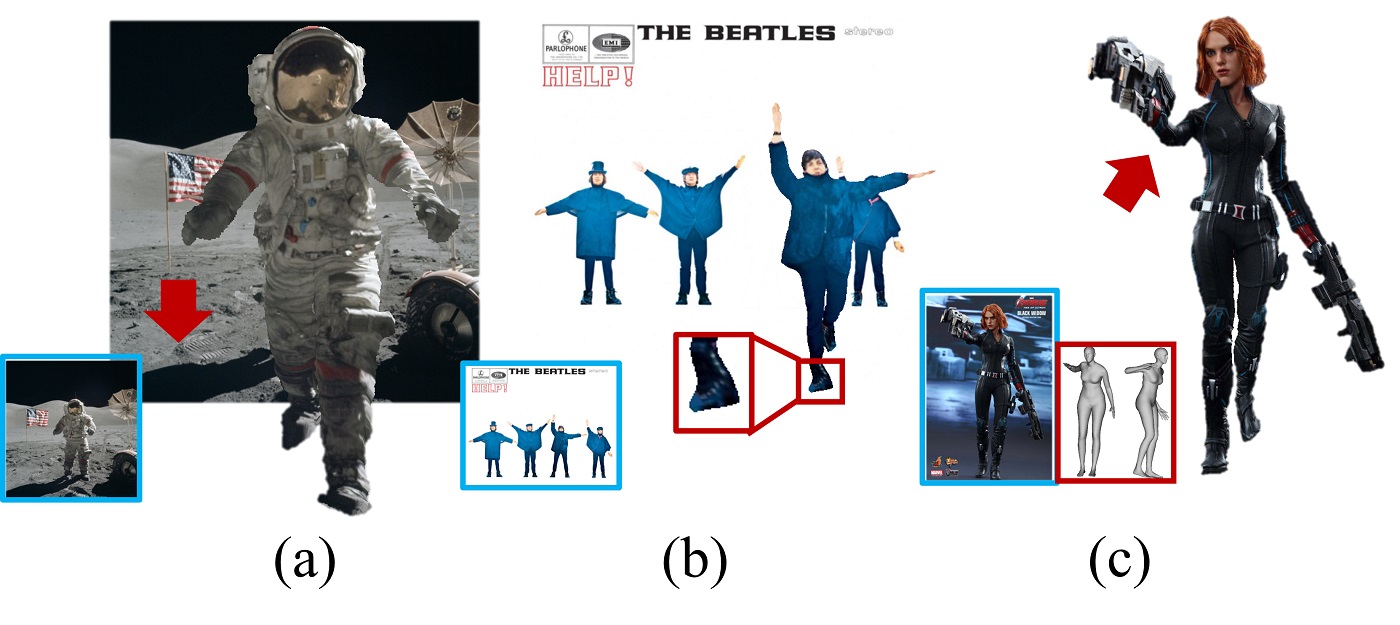}
  \caption{Examples of limitations (inputs in blue boxes). (a) Shadows not modeled. (b) Unrealistic mesh. (c) 3D pose error leads to incorrect geometry.} 
  \label{fig:limitation}
\end{figure}

\textbf{Limitations:} We note the following limitations (see also Fig.~\ref{fig:limitation}): (1) Shadows and reflections are currently not modeled by our method and thus won't move with the animation. (2) The SMPL model may produce incorrect 3D pose due to ambiguities. (3) Since the reconstructed mesh must fit the silhouette, the shape may look unrealistic, e.g., wrong shape of shoes; on the other hand this enables handling abstract art. (4) Our method accounts for self-occlusions when arms partially occlude the head, torso, or legs. It remains future work to handle other occlusions, e.g., legs crossed when sitting. (5) We have opted for simple texture inpainting for occluded body parts, with some user interaction if needed.  Using deep learning to synthesize, e.g., the appearance of the back of a person given the front, is a promising research area, but current methods that we have tested~\cite{ma2017pose, esser2018variational, ma2017disentangled} give very blurry results.

{\bf Summary:} We have presented a method to create a 3D animation of a person in a single image. Our method works with large variety of of whole-body, fairly frontal photos, ranging from sports photos, to art, and posters. In addition, the user is given the ability to edit the human in the image, view the reconstruction in 3D, and explore it in AR. 

We believe the method not only enables new ways for people to enjoy and interact with photos, but also suggests a pathway to reconstructing a virtual avatar from a single image while providing insight into the state of the art of human modeling from a single photo.

{\bf Acknowledgements} The authors thank Konstantinos Rematas for helpful discussions, Bogo et al. and Alldieck et al. for sharing their research code, and labmates from UW GRAIL lab for the greatest support. This work was supported by NSF/Intel Visual and Experimental Computing Award \#1538618, the UW Reality Lab, Facebook, Google, Huawei, and a Reality Lab Huawei Fellowship.


{\small
\bibliographystyle{ieee}
\bibliography{egbib}
}

\end{document}